\documentclass[conference]{IEEEtran}
\pdfoutput=1
\usepackage{cite}
\usepackage[pdftex]{graphicx}
\graphicspath{{/images/}}
\DeclareGraphicsExtensions{.jpeg,.png}
\usepackage{amsmath}
\usepackage{algorithm}
\usepackage{algpseudocode}
\usepackage{array}
\usepackage[caption=false,font=footnotesize]{subfig}
\usepackage[utf8x]{inputenc}
\usepackage{url}
\usepackage{amsfonts}
\usepackage{color,soul}

\begin{document}

\title{Optimal Experimental Design of Field Trials \\ using Differential Evolution \\ {\large An application in Quantitative Genetics and Plant Breeding}}

\author{
\IEEEauthorblockN{Vitaliy Feoktistov}
\IEEEauthorblockN{Stéphane Pietravalle}
\IEEEauthorblockN{Nicolas Heslot}
\IEEEauthorblockA{
Biostatistics Department \\
Research Centre of Limagrain Europe \\
Chappes, France \\
vitaliy.feoktistov@limagrain.com}
}

\maketitle

\begin{abstract}
When setting up field experiments, to test and compare a range of genotypes (e.g. maize hybrids), it is important to account for any possible field effect that may otherwise bias performance estimates of genotypes. To do so, we propose a model-based method aimed at optimizing the allocation of the tested genotypes and \textit{checks} between fields and placement within field, according to their kinship. This task can be formulated as a combinatorial permutation-based problem. We used Differential Evolution concept to solve this problem. We then present results of optimal strategies for between-field and within-field placements of genotypes and compare them to existing optimization strategies, both in terms of convergence time and result quality. The new algorithm gives promising results in terms of convergence and search space exploration.
\end{abstract}


% \fi
%\IEEEpeerreviewmaketitle

\section{Introduction}\label{sec:intro}

In plant breeding trials, field experiments are commonly used to test genotypes (e.g. maize hybrids or wheat varieties) and estimate their potential for a range of variables, such as yield, at all stages of the breeding process. Accurately estimating those variables is crucial to ensure that only the best genotypes are kept for further selection.

In a field experiment, each genotype is allocated to one or several small plots on a field. Measured performance on that plot is due to an effect of the genotype, measurement error and an effect of the field (e.g. nutrient availability). Fisher highlighted the importance of experimental design in the early 1930s \cite{Fisher1935}, to reduce bias and better estimate effects of interest.

In its most basic form, experimental design rely on randomization and replication but better design structures, for instance, through the use of blocking, can further improve the reliability of the estimates. However, blocking structures rely on the strong assumption of local homogeneity and also require the design to be replicated. As a result, they can sometimes be difficult to use, especially in early generation variety trials, where the aim is to test a very large number of genotypes with constraints on space and seed availability. In recent years, model-based designs have been used more widely \cite{Gilmour1997,Cullis2006,ButlerDavidG.SmithAlisonB.2014}. At the local scale, one example of such a design is the repchecks design and relies on a spatial correlation structure of error first assumed by Gilmour et al. \cite{Gilmour1997}. This design consists of a number of unreplicated \textit{experimental} genotypes and some (usually three or four) replicated and well spread-out \textit{check} genotypes to capture and model any potential field effect. In plant breeding trials, it is important to test the genotypes over a range of locations to estimate their performance over a network (e.g. market). Similarly to the constraints present locally, it is often practically impossible to have all genotypes replicated over all locations. As a result, many genotypes are only present on a subset of the locations in the network; this can be achieved through multi-location designs such as the sparse repchecks or sparse p-reps designs \cite{Williams2014a, Williams2011b}.

The Section~\ref{sec:problem} describes the optimal design of field experiments problem. Its mathematical formulation as well as modelling as combinatorial permutation-based optimization task. The problem is split into two phases to be solved efficiently. The Section~\ref{sec:de} describes a general framework of Differential Evolution (DE) algorithm, and then introduces the new created algorithm for exploring the permutation space. This is followed by proposition of several search strategies, it also mentions the used method to handle constraints. The Section~\ref{sec:results} demonstrates the obtained results for both optimizations. Here we show the efficiency of the approach and compare it with the existing software packages. Real-life examples are presented. The Section~\ref{sec:conclusion} concludes the paper outlining promising trends and directions.

\section{Problem Formulation}\label{sec:problem}

\subsection{State of the Arts and Motivation}

Two PhD theses tried to tackle this problem from a practical viewpoint \cite{Coombes2002,Butler2013}. That resulted in two software packages: DiGGer \cite{Coombes2009}, a design search tool in R \cite{RDevelopmentCoreTeam2016}, and OD \cite{Butler2014}, also an R-package. Other recent papers contribute to improving the mathematical model of efficient designs \cite{ButlerDavidG.SmithAlisonB.2014, Williams2014a, Williams2011b} and show real-case studies.

The core of both software packages is coded in Fortran and highly optimized. The problem can be considered as a HPC one and despite of all code tunings, the computations are still very expensive. To solve this problem, many metaheuristics have been tested, among them: Tabu Search, Simulated Annealing, Genetic Algorithms, and others. Nevertheless, for many real cases the mentioned algorithms are not suitable due to their computational time, in other words a poor convergence. This motivated us to develop a new approach to solve this problem more efficiently.

\subsection{Modelling Optimal Design as a Permutation Problem}
Let us consider the following linear mixed model \cite{Searle1992}

\begin{equation}
y=X\beta + Zu + \varepsilon
\end{equation}
$y$ – is vector of phenotype, $X$ – a design matrix for the fixed effects, $\beta$ – are fixed effects, $Z$ – a design matrix for the random genetic effect $u$, so that $u$ is distributed multivariate normal with covariance $G$, and $\varepsilon$ – the error term is multivariate normal with covariance $R$.  $G=K\sigma_a^2$, where $\sigma_a^2$  – is the additive genetic variance and $K$ is the kinship which can be based on pedigree or on markers. Kinship information can be ignored in the optimization process by setting $K$ to an identity matrix.

From this model, the mixed model equations can be written \cite{Henderson1984br}
\begin{equation}
\begin{bmatrix}
X^{'} R^{-1} X & X^{'} R^{-1} Z \\
Z^{'} R^{-1} X & Z^{'} R^{-1} Z + G^{-1}
\end{bmatrix}
\begin{pmatrix}
	\beta \\ u
\end{pmatrix}
=
\begin{pmatrix}
X^{'} R^{-1} y \\
Z^{'} R^{-1} y
\end{pmatrix}
\end{equation}

In a general case, the prediction error variance PEV is
\begin{equation}\label{eq:PEV}
PEV=var(u-\hat{u} )=[Z^{'}MZ+G^{-1} ]^{-1}
\end{equation}
where
\begin{equation}\label{eq:M}
M=R^{-1}-R^{-1} X(X^{'} R^{-1} X)^{-1} X^{'} R^{-1}
\end{equation}

For more details, see \cite{Henderson1984e} and \cite{Laloe1993}.

The optimization task consists in finding an optimal design matrix $Z^*$, which minimizes the $PEV$. Let exists some start design matrix $Z_0$  with given properties, then the optimization task can be rewritten as finding the optimal permutation $\pi^*$, so that $Z^*=\pi^*(Z_0)$ is the permutation of lines of $Z_0$ matrix with respect to constraints representing desirable design properties.

Taking into account relatedness between genotypes ensures an optimal design is generated avoiding planting closely related genotypes next to each other and splitting families evenly between locations. This is a $NP$-hard problem and is hard to solve for large sized cases, i.e. real situations. Thus, the task is split into two phases. The first one (phase I) is called \textit{Between Location} phase, and optimizes dispatching the genotypes among several locations (sparse design). The second one (phase II) is \textit{Within Location} phase. At this stage, the optimal design inside of a location is resolved taking in consideration auto-regressive error and genotype subsets from the previous phase. As shown in Eqs.~\ref{eq:PEV} and ~\ref{eq:M}, for both phases, the optimization can also take into account additional fixed effects estimation such as block effects within location.

The incorporation of a spatial auto-regressive process $(ar(1) \times ar(1))$  with a so-called measurement error variance ($nugget$) implies a given structure in the variance-covariance matrix of the residuals of the mixed model. Assuming a rectangular field of size $r$ rows and $c$ columns, allowing the last row to be incomplete (with $c_{l}$ columns), the matrix $R$ is of size $[c\cdot(r-1)+c_{l},\;c\cdot(r-1)+c_{l}]$ and can be defined as
\begin{equation}\label{eqn:R_matrix}
\begin{array}{rcl}
R_{i,i}& = & \psi \\
R_{i,j}& = & \rho^{\left| a_{i}-a_{j} \right| }_{r} \cdot \rho^{\left| b_{i}-b_{j} \right| }_{c}
\end{array}
\end{equation}
where $\rho_{r}$ and $\rho_{c}$ are the autocorrelation coefficients for the rows and columns respectively, $\psi=1+nugget$ and
\begin{equation}
\begin{array}{ll}
a_{i}&=(i-1)\;\mbox{mod}\;c + 1 \\
a_{j}&=(j-1)\;\mbox{mod}\;c + 1 \\
b_{i}&=(i-1)\;\mbox{quo}\;c + 1 \\
b_{j}&=(j-1)\;\mbox{quo}\;c + 1 
\end{array}
\end{equation}

\section{Differential Evolution}\label{sec:de}

\subsection{General Description}
Since its first introduction in 1995 \cite{Storn1995b, Storn1995} Differential Evolution is one of the most successful metaheuristic algorithm. It has won many CEC global optimization competitions. The algorithm was generalized to different branches of optimization such as constrained, multi-modal, multi-objective and large-scale optimization as well as made suitable for noisy and mixed-variables objective functions. Thus it covers with success many applications in a number of areas of science and industry. A detailed history of the algorithm can be found in Chapter 1 of \cite{Feoktistov2006}.

Differential Evolution is a population-based approach. Its concept shares the common principles of evolutionary algorithms. Starting from an initial population $\mathcal{P}_0$ of $NP$, solutions DE intensively progresses to the global optimum using self-learning principles. The initialization can be done in different ways, the most often uniformly random solutions are sampled respecting boundary constraints. Then, each solution is evaluated through the objective function.

To be formal, let the population $\mathcal{P}$ consists of $ind_i: i=1,NP$ solutions, or \textit{individuals} in evolutionary algorithms terms. The individual $ind_i$ contains $D$ variables $x_{ij}$, so-called \textit{genes}. Thus $ind_i=\{x_{ij}\}_{j=1}^D$ and $\mathcal{P}=\{ind_i\}_{i=1}^{NP}$.

At the each iteration $g$, also called \textit{generation}, all individuals are affected by reproduction cycle. To this end, for each current individual $ind$ a set of other individuals $\varTheta = \{ \xi_1, \xi_2,\ldots, \xi_n\}$ are randomly sampled from the population $\mathcal{P}_g$. All the DE search strategies are designed on the basis of this set.

The \textit{strategy} \cite{Feoktistov2004} is how the information about $\varTheta$ set individuals is used to create the base $\beta$ and the difference $\delta$ vectors of the main DE operator of the reproduction cycle, often called as differential mutation by analogy with genetic algorithms, or else \textit{differentiation} by functional analogy with gradient optimization methods. So the differentiation operator can be now viewed in its standardized form as the current point $\beta$ and the stochastic gradient step $F\delta$ to do
\begin{equation}\label{eq:differentiation}
\omega = \beta + F \cdot \delta
\end{equation}
where $F$ is \textit{constant of differentiation}, one of the control parameters. Usually the recommended values are in $[0.5,2)$ range.

The next reproduction operator is \textit{crossover}. It is a typical representative of genetic algorithms' crossovers. Its main function is to be conservative when passing to the new solution preserving some part of genes from the old one. The most used case is when the \textit{trial} individual $\omega$ inherits the genes of the target one with some probability
\begin{equation}
	\omega_j = \left\lbrace
	\begin{array}{ll}
		\omega_j & \mbox{if} \quad rand_j \geq Cr \\ 
		ind_j & \mbox{otherwise}
	\end{array} \right.
\end{equation}
for $j=1,\ldots,D$, uniform random numbers $rand_j \in [0,1)$, and \textit{crossover constant} $Cr \in [0,1)$, the second control parameter.

So, the trial individual $\omega$ is formed and the next step is to evaluate it through an objective function, also called \textit{fitness}. Sometimes, constraints are handled at this stage too. Reparation methods are used before the fitness evaluation, in order to respect the search domain $[L,H]$ and/or $c(\omega) \leq 0$ (hard approach). Other constraints can be evaluated during the objective function computation (soft approach). It should be noted that in many industrial applications the evaluation of objective function and constraints demands significant computational efforts in comparison to other parts of the DE algorithm, thus the algorithm's convergence speed during the first iterations is very important.

The following step is \textit{selection}. Often, the \textit{elitist} selection is preferred

\begin{equation}
	ind = \left\lbrace
		\begin{array}{ll}
			\omega & \mbox{if} \quad f(\omega) \leq f(ind) \\ 
			ind & \mbox{otherwise}
\end{array} \right.
\end{equation}
At this moment the decisions on constraints and multi-objectives values are also influence the final choice of the candidate.

The Differential Evolution pseudo-code is summarized in Alg.~\ref{alg:de}.

\begin{algorithm}
	\caption{Differential Evolution - a general pattern}\label{alg:de}
	\begin{algorithmic}[0]
		\Statex\textbf{Require} $params$ :
		\State $\quad F,Cr,NP,\text{Strategy}$ \Comment{control parameters}
		\State $\quad f(\cdotp)$ \Comment{objective function}
		\State $\quad c(\cdotp), L, H$ \Comment{constraints}
		\Procedure {DifferentialEvolution}{$params$}
		\State Initialize $\mathcal{P}_0 \gets \{ind_1,\ldots,ind_{NP}\} \in [L,H]$
		\State Evaluate $f(\mathcal{P}_0) \gets \{f(ind_1), \ldots, f(ind_{NP})\}$
		\While{not stopping condition}
			\ForAll {$ind \in \mathcal{P}_g$}
				\State Sample $\varTheta = \{ \xi_1, \xi_2,\ldots, \xi_n\}$ from $\mathcal{P}_g$
				\State Reproduce
				\State $\quad \omega \gets Differentiation(\varTheta,F,\text{Strategy})$
				\State $\quad \omega \gets Crossover(\omega,ind,Cr)$
				\State Evaluate $f_{\omega} \gets f(\omega)$ \Comment{often costly operation}
				\State Select $(\omega$ vs $ind) \rightarrow ind$ 
			\EndFor
			\State $g \gets g + 1$ \Comment{go to the next generation}
		\EndWhile
		\EndProcedure
	\end{algorithmic}
\end{algorithm}

\subsection{Adaptation to the Permutation Space}

Differential Evolution is very successful in solving continuous optimization problems. Nevertheless, there were many attempts to spread the DE concept on combinatorial optimization. A good summary of DE adaptations can be found in \cite{GodfreyOnwuboluBook2009}. In this paper, we concentrate on the combinatorial problems than can be formulated as permutation-based. We propose a new technique to explore combinatorial space and apply this technique to find the optimal design of field experiments, a real application from agriculture and plant breeding. 

Further we discuss how to transform the differentiation operator to handle the permutation space. There are two points we need to define
\begin{enumerate}
	\item what would be the distance  $\delta$ in permutation space? and
	\item how we apply this distance knowledge to compute the DE step from the base point $\beta$ (see Eq.~\ref{eq:differentiation})?
\end{enumerate}

As it was shown earlier \cite{Feoktistov2006}, the crossover operator is optional and can be missed in many cases without degradation of convergence rate. So we decided not to use it for a permutation-based optimization.

\subsubsection{Distance}
Many types of distances for the permutation space were invented, explored and used. A good overview of combinatorial distances with indicating their complexity and performance tests is presented in \cite{Zaefferer2014}.

It is obvious that the Hamming distance is one of the best candidates because of its simplicity to compute $\mathcal{O}(n)$ and suitability for the DE context. For two permutations $\pi$ and $\pi'$, the Hamming distance between them is
\begin{equation}\label{eq:distance}
	\varDelta_\mathcal{H} (\pi,\pi') = \sum_{j=1}^{D} \chi_j 
	\quad\text{where } \chi_j = \left\lbrace
			\begin{array}{ll}
				1 & \mbox{if} \quad \pi_j \neq \pi'_j \\ 
				0 & \mbox{otherwise}
			\end{array} \right.
\end{equation}

\subsubsection{DE step}
Several combinatorial operators are possible to generate the permutation $\pi$. Most common are swap, interchange and shift. Swap is considered a local operator as it concerns the neighbours of some permutation position $\pi_j$. Interchange is a global operator, it swaps two distant positions $\pi_i$ and $\pi_j$. Shift interchanges two $k$-length substrings $[\pi_{i},\ldots,\pi_{i+k}]$ and $[\pi_{j},\ldots,\pi_{j+k}]$.

Many combinatorial heuristic algorithms alternate proportions of local and global actions to balance the search process in order to avoid stagnation or unnecessary exploration. We do not use the different operators to trade-off local and global techniques but chose only one global operator, namely \textit{interchange} and control it with \textit{locality} factor $\lambda$, which can be considered as an equivalent of the $F$ differentiation constant but for the permutation space.

Thereby the size of DE steps is defined as $\lambda\varDelta_\mathcal{H}(\delta)$. Thus the trial individual is computed as
\begin{equation}
\omega = \beta \oplus \lambda \varDelta_\mathcal{H}
\end{equation}
where the operator $\oplus$ is an interchange operator, which applies $\lambda \varDelta_\mathcal{H}$ successive interchanges on the $\beta$ vector.

It should be noted that this is a general concept and any distance metric $\varDelta$ and any permutation operator $\odot$ may be used instead of these ones. Their right combination depends mainly on the problem to solve. Also, the locality factor $\lambda$ may be self-learned or real-time adjusted.

\subsection{Some Strategies and their Properties}

The strategies are key to the search space exploration. They propose different techniques on how to choose and to construct $\beta$ and $\delta$ from the set of sampled individuals $\varTheta$. Here we give three examples having different search properties.

\subsubsection{rand3}
For each individual $ind$ a set $\varTheta$ of three other random individuals $\{\xi_1, \xi_2,\xi_3\}$ is sampled from the population $\mathcal{P}_g$. The vector $\delta$ is defined by metric on $\xi_1$ and $\xi_2$, that is $\delta(\xi_1,\xi_2)=\lVert \xi_1 - \xi_2 \rVert_{\varDelta_\mathcal{H}}$, in other words, the distance is calculated as $\varDelta_\mathcal{H} (\xi_1, \xi_2)$ (see Eq.~\ref{eq:distance}). $\beta = \xi_3$. So the trial
\begin{equation}
\omega = \xi_3 \oplus \lambda \cdot \lVert \xi_1 - \xi_2 \rVert_{\varDelta_\mathcal{H}}
\end{equation}

This strategy is good for most cases, especially for large-scale problems when an intensive exploration is needed. When $\lambda \rightarrow D$, the strategy turns into a random search.

\subsubsection{rand2best}
Let $best$ is the current best individual, that is $f(best) \leq f(ind_i) \quad \forall ind_i \in \mathcal{P}_g$. Two additional random individuals are extracted, $\varTheta=\{\xi_1,\xi_2\}$, to compute $\delta(\xi_1,\xi_2)$. $\beta=best$. Thus
\begin{equation}
\omega = best \oplus \lambda \cdot \lVert \xi_1-\xi_2 \rVert_{\varDelta_\mathcal{H}}
\end{equation}

This strategy is inspired from social behaviour, when there is an alternated leader organizing and directing the others, and others have a tendency to be attracted by the leader. It also has some analogies with Particle Swarm Optimization \cite{PSO1995}. This strategy is efficient for mid-sized problems. It provides a good balance between exploration and exploitation.

\subsubsection{dir2best}
Here, the $best$ individual and two others randomly extracted are selected, but this time the individuals are ordered as follow: $f(\xi_1) \leq f(\xi_2)$ and $best \neq \xi_1$. $\beta=\xi_1$, so
\begin{equation}
\omega = \xi_1 \oplus \lambda \cdot \lVert \xi_2-best \rVert_{\varDelta_\mathcal{H}}
\end{equation}

This strategy uses stochastic gradient information, that is the distance to the $best$ individual. It has excellent convergence properties on small- and mid- size problems. In large-scale tasks, because of an intensive exploitation, it has a tendency to premature convergence (Chapter $4$ of~\cite{Feoktistov2006}).

\subsection{Constraints}\label{sec:Constraints}

The initial matrix $Z_0$ (see Eq.~\ref{eq:PEV}) possesses some design properties to keep. This represents the set of constraints to respect $c(\omega) \leq 0$. Thereby a hard approach for constraints handling is most suitable. We apply it at the stage of the differential operator. As the objective function is long to compute, it is easier and more efficient is to evaluate only feasible individual. This way, we significantly reduce the space of exploration without doing unnecessary evaluations.

\section{Results}\label{sec:results}

\subsection{Phase I – Between Locations}\label{sub:Between}

In this section, we present the results of the allocation of genotypes across locations. To illustrate the impact of accounting for the kinship matrix $K$, we use an extreme case of a population of $403$ genotypes and five locations (of $300$ plots each). The $403$ genotypes are split into three \textit{check} genotypes (present and replicated $20$ times on each location) and $400$ \textit{experimental} genotypes. We further impose for these $400$ genotypes to each be present once on three out of the five locations. We then define them as belonging to three independent families (sizes $14$, $187$ and $199$) of full siblings, by using a block matrix for the kinship matrix $K$ calculated using pedigree. 

\begin{equation}
K
=\begin{bmatrix}
A_{14} & 0 & 0 \\
0 & B_{187} & 0 \\
0 & 0 & C_{199}
\end{bmatrix}
\end{equation}
where $A$, $B$ and $C$ are square matrices whose off-diagonal elements are $0.5$ and diagonal elements are $1$.

Following the constraints imposed on them, it is clear that the \textit{check} genotypes can be excluded from this first optimization phase. Because of the small size of family $1$, a random allocation of genotypes across locations can lead to a very unbalanced spread as shown in Table~\ref{tab:Phase I} below, where this family is under-represented in location $3$, therefore causing a risk of biasing the corresponding genotypes estimates (confounding between location and family effects).

\begin{table}[h]
	\caption{\label{tab:Phase I}Phase I - Initial family spread across locations}
	\centering
	\begin{tabular}{l||c|c|c}
		& \textbf{Family 1} & \textbf{Family 2} & \textbf{Family 3} \\ 
		\hline \hline
		\textit{Location 1} & 9 & 111 & 120 \\ 
		\textit{Location 2} & 9 & 116 & 115 \\ 
		\textit{Location 3} & 1 & 116 & 123 \\ 
		\textit{Location 4} & 13 & 110 & 117 \\
		\textit{Location 5} & 10 & 108 & 122 \\ 
	\end{tabular}	
\end{table}

Here, we therefore have, using Eq.~\ref{eq:PEV} and Eq.~\ref{eq:M},
\begin{equation}\label{eqn:eqlabel}
\begin{array}{ll}
R & = \mathbf{I}_{1200} \\
X & = \begin{bmatrix}
\mathbf{1}_{240} & 0 & 0 & 0 & 0\\
0 & \mathbf{1}_{240} & 0 & 0 & 0\\
0 & 0 & \mathbf{1}_{240} & 0 & 0\\
0 & 0 & 0 & \mathbf{1}_{240} & 0\\
0 & 0 & 0 & 0 & \mathbf{1}_{240}\\
\end{bmatrix}
\end{array}
\end{equation}
and $Z_{0}$ is the design matrix, size $[1200, 400]$.

\begin{figure}[h]
	\centering
	\includegraphics[width=0.9\linewidth]{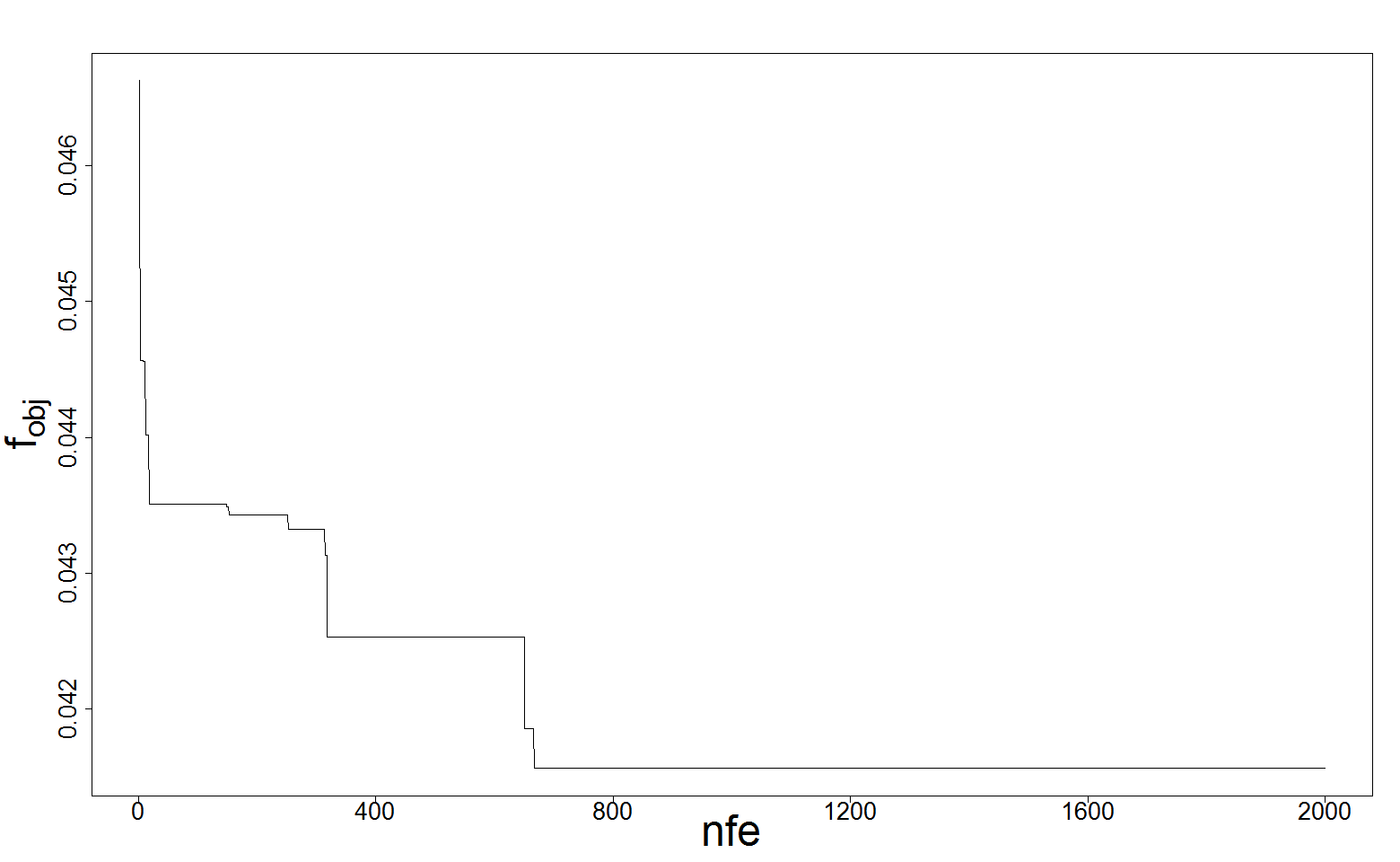}
	\caption[Optimization function]{Phase I - Convergence rate: \textit{nfe} - number of function evaluations.}
	\label{fig:Phase I}
\end{figure}

Fig.~\ref{fig:Phase I} shows the convergence of the objective function for this first phase. This was obtained using $30$ restarts, each with $2000$ steps (number of function evaluations) and convergence was reached in $383.1$ seconds. The objective function was approximated by single value decomposition, using the first three eigenvalues, the population size $NP$ was set at $25$ and the search strategy used was $rand3$. For each simulation, we allocated only six cores (threads) of a Intel Xeon E5-4627 v3 processor under Windows Server 2012 R2 operating system.

The convergence value of the process $(0.04156354)$ shows a good improvement over the objective function value of the design $(0.04841361)$ randomly generated. This can be seen in Table~\ref{tab:Phase I - Optimum}, where the spread of each family across locations is better balanced.

However, as mentioned in Section~\ref{sec:intro}, experimental designs can have many constraints imposed on them, depending on their complexity. Further work is ongoing to add additional constraints (e.g. repetition of \textit{experimental} genotypes within location).

\begin{table}[h]
	\caption{\label{tab:Phase I - Optimum}Phase I - Optimized strategy}
	\centering
	\begin{tabular}{l||c|c|c}
		& \textbf{Family 1} & \textbf{Family 2} & \textbf{Family 3} \\ 
		\hline \hline
		\textit{Location 1} &  9 & 121 & 110 \\ 
		\textit{Location 2} & 13 & 105 & 122 \\ 
		\textit{Location 3} &  7 & 117 & 116 \\ 
		\textit{Location 4} &  7 & 107 & 126 \\ 
		\textit{Location 5} &  6 & 111 & 123 \\
	\end{tabular}
\end{table}

\subsection{Phase II – Within Location}

After running the optimization across locations, it is necessary to run this phase independently over each of the locations, after having re-incorporated the \textit{check} genotypes, to obtain an optimal field layout.

The first part of this section looks at the gain made through this approach, compared to an approach currently used widely, DiGGer. In the second part of this section, we investigate the effect of using the kinship information in the local (i.e. within location) optimization. To do so and compare this algorithm to the existing algorithm (DiGGer), we use an extreme case of a field with $144$ plots in twelve rows and columns where we aim at allocating a repchecks design with $119$ \textit{experimental} genotypes and three \textit{check} genotypes, where the three \textit{checks} are repeated nine, eight and eight times respectively. Further, we define the $122$ genotypes as belonging to three independent families of size $40$, $40$ and $42$ respectively, each with one of the \textit{check} genotypes. In both examples presented below, we used $6$ restarts, each with $10000$ steps (number of function evaluations), the population size $NP$ was set at $25$ and the search strategy used was $rand2best$.\\

\subsubsection{Ignoring kinship}
As well as defining the $X$ and $Z$ matrices in the same way as during the between locations optimization, it is important to account for the local autocorrelation defined by $R$. The matrix $R$ used in this case is derived using autocorrelation coefficients $\rho_{r}$ and $\rho_{c}$ of $0.5$, based on data collected on existing trials. We further assume a trait with a large $(h^{2}=0.8)$ heritability (analogous to the noise/signal ratio) to ensure that a good spatial spread of the genotypes is critical during the trial set-up. To allow a direct comparison between the method proposed here and DiGGer, we define the kinship matrix $K$ as the identity matrix, therefore ignoring relatedness between genotypes.

\begin{figure}[h]
	\centering
	\includegraphics[width=0.9\linewidth]{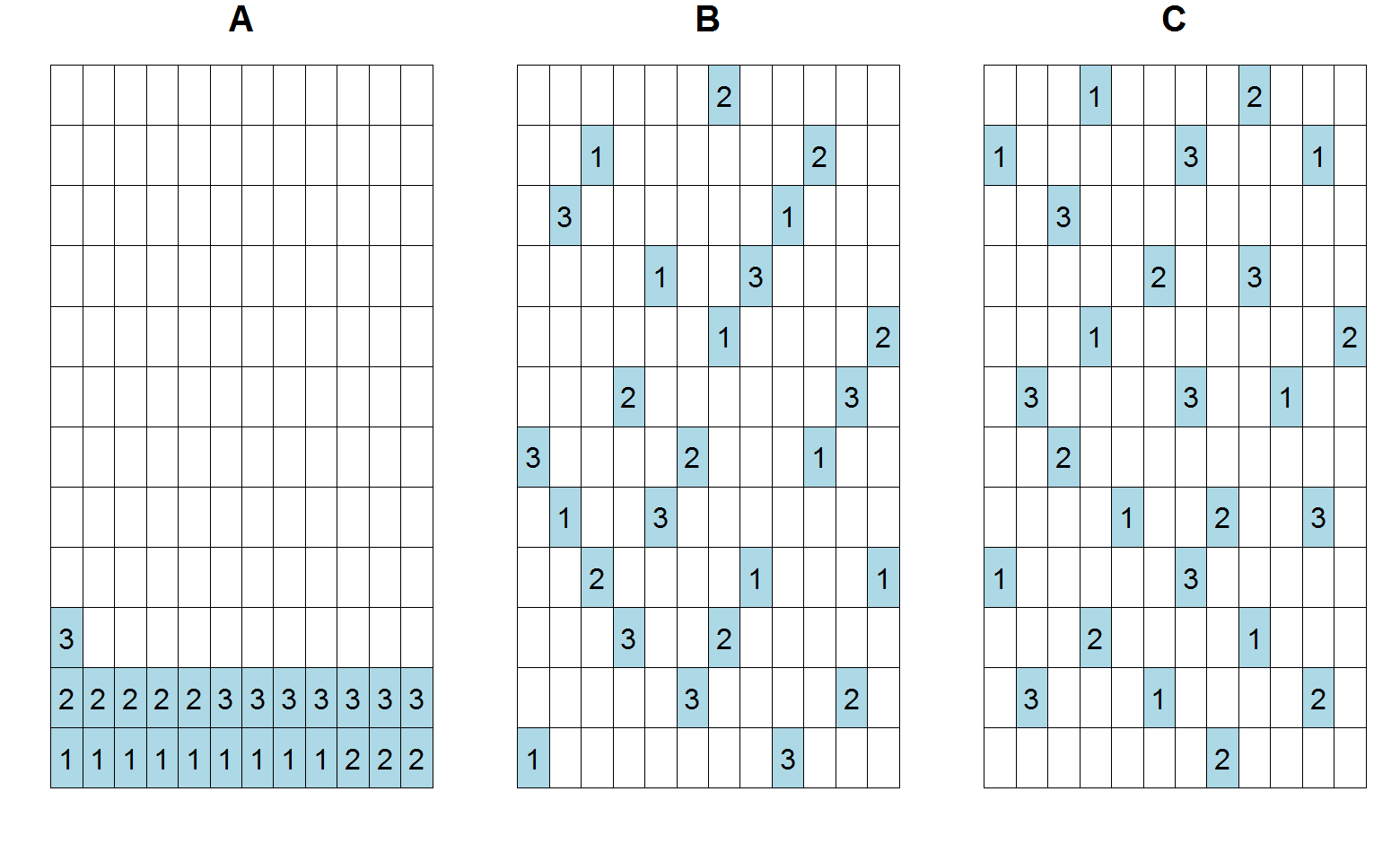}
	\caption[Optimization function]{Phase II - Allocation: (A) \textit{Checks} at start; (B) \textit{Checks} optimized - DiGGer;	(C) \textit{Checks} optimized - New Algorithm}
	\label{fig:Phase II}
\end{figure}

\begin{figure}[h]
	\centering
	\includegraphics[width=0.9\linewidth]{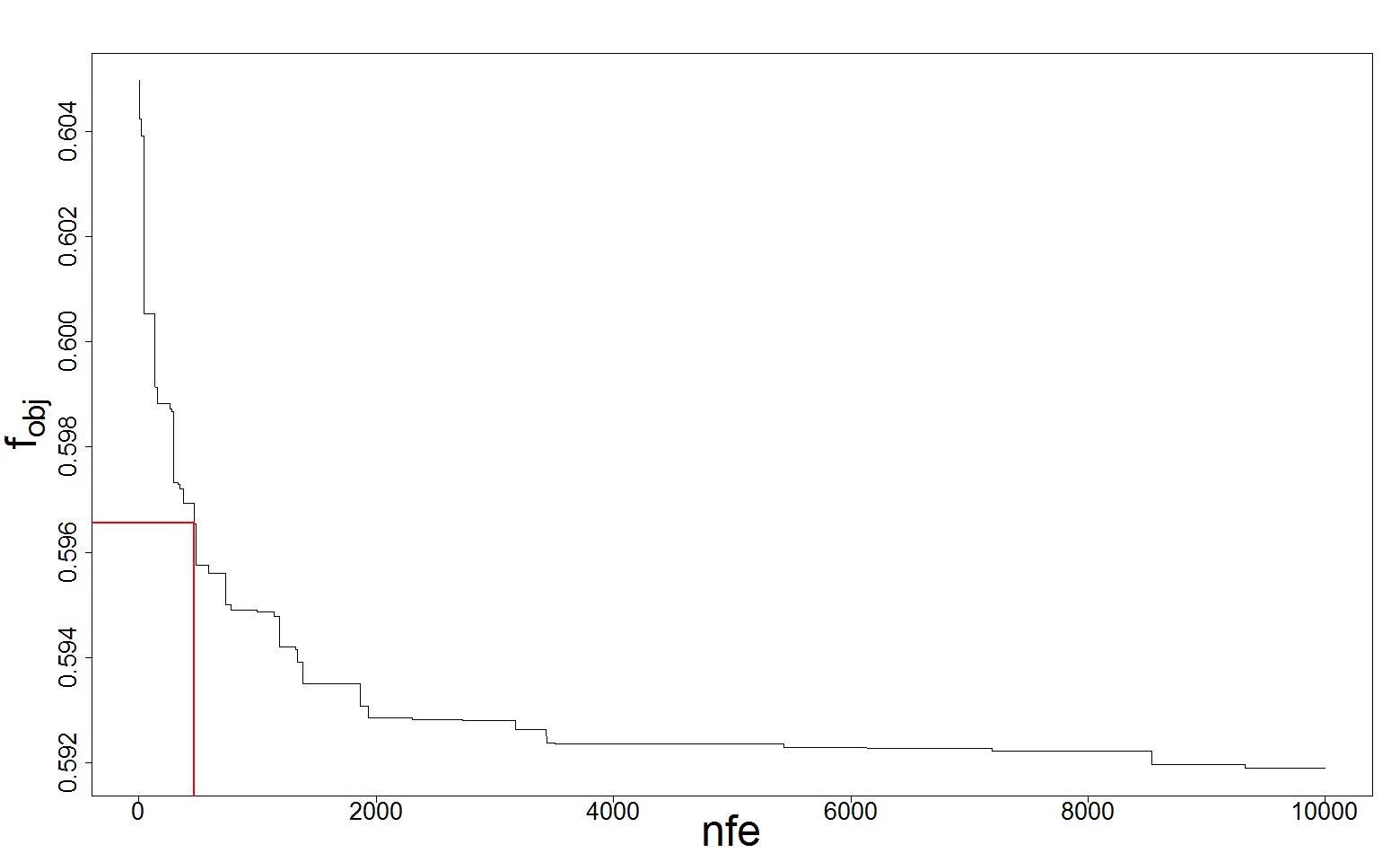}
	\caption[Optimization function]{Phase II - Convergence rate: \textit{nfe} - number of function evaluations.\\		
		The red lines correspond to the value of the objective function for the design as optimized by DiGGer and the number of functions needed for our algorithm to better it}
	\label{fig:Phase II-Cv}
\end{figure}

In Fig.~\ref{fig:Phase II}, we present an example of a field within a location, where each small rectangle is a plot and corresponds to a genotype. The first figure shows the starting design, where we purposely grouped all three \textit{check} genotypes at one end of the field, to see the impact of the optimization. The middle figure shows the spatial spread of those after the optimization procedure derived using DiGGer. Unsurprisingly, because of the equal weight given to both dimensions in the autoregressive process in DiGGer, the optimized strategy shows clear diagonals of \textit{check} genotypes. Further, a common feature of DiGGer can also be observed: those diagonals are often made of adjacent plots (two diagonals length four in the example presented). The third plot of the figure presents the optimization obtained through our algorithm and shows a visually better spread of the \textit{check} genotypes; although the pattern of diagonals is still present, it is better spread out throughout the field, with no more than two adjacent plots. Fig.~\ref{fig:Phase II-Cv} presents the convergence speed of the algorithm, together with the value of the objective function reached by the DiGGer algorithm. In this example, convergence was reached in $24.4$ seconds using our algorithm and in $9.9$ seconds using DiGGer. Further, note that, unlike our algorithm, DiGGer is mono-thread and cannot use the benefit of multi-core servers.

The visually better design is confirmed when looking at the value of the objective function produced by DiGGer $(0.59656328)$ and the new algorithm $(0.59188730)$. However, it is important to note that the algorithm used in DiGGer is based on the Reactive Tabu Search \cite{Coombes2002}. Note that, in this case, using only $2000$ steps, whilst keeping everything else identical, led to a convergence time of $5.3$ seconds without dramatically affecting the value of the objective function ($0.59340239$, i.e. still improving on the convergence value produced by DiGGer).

\subsubsection{Accounting for kinship}
In this second case, we include, in the optimization phase, a non-identity kinship matrix $K$. The matrices $X$ and $Z_{0}$ are defined as in \ref{sub:Between}. Further, the 122 genotypes are split in three families of full siblings:

\begin{equation}
K
=\begin{bmatrix}
A_{40} & 0 & 0 \\
0 & B_{40} & 0 \\
0 & 0 & C_{42}
\end{bmatrix}
\end{equation}
where $A$, $B$ and $C$ are square matrices whose off-diagonal elements are 0.5 and diagonal elements are $1$.

\begin{figure}[h]
	\centering
	\includegraphics[width=0.9\linewidth]{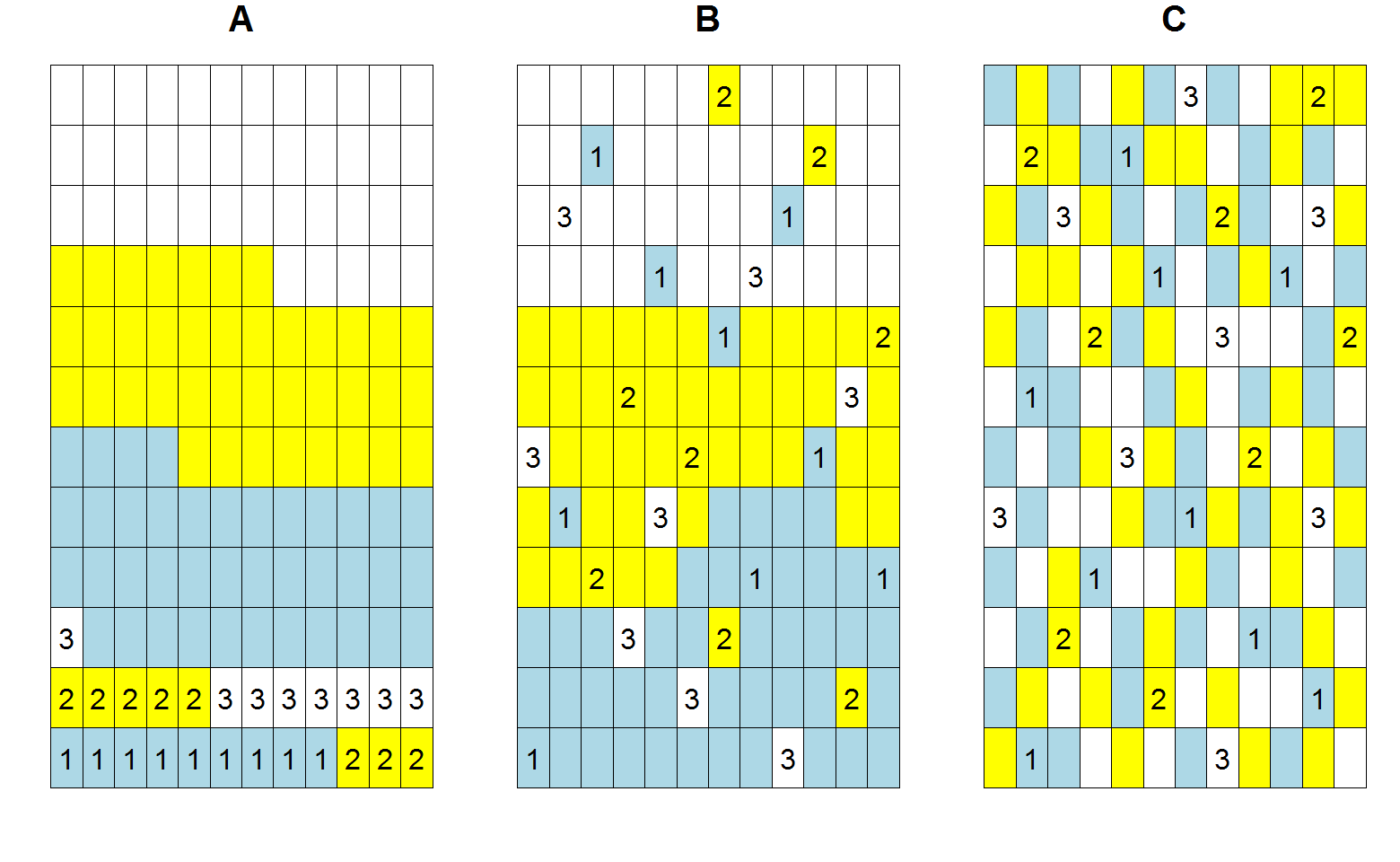}
	\caption[Optimization function]{Phase II - Allocation using kinship information: (A) - Families at start; (B) - Families optimized using DiGGer; (C) - Families optimized using the New Algorithm.}
	\label{fig:Phase II with A}
\end{figure}

\begin{figure}[h]
	\centering
	\includegraphics[width=0.9\linewidth]{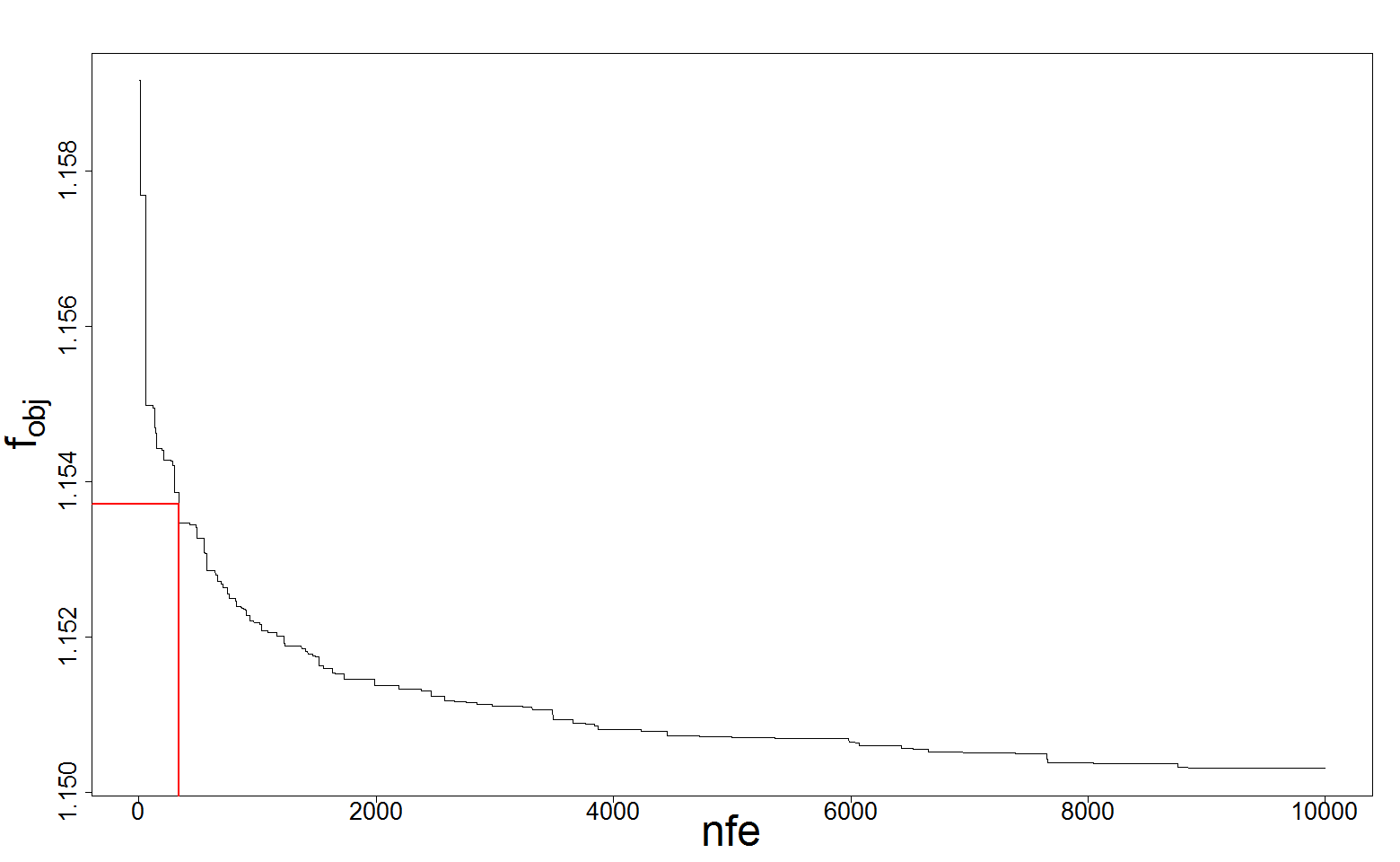}
	\caption[Optimization function]{Phase II (incl. kinship information) - Convergence rate: \textit{nfe} - number of function evaluations.\\
		The red lines correspond to the value of the objective function for the design as optimized by DiGGer, when the \textit{experimental} genotypes are randomized at start, and the number of functions needed for our algorithm to better it}
	\label{fig:Phase II cv with kinship}
\end{figure}

Fig 4 shows the initial design, the design as optimised by DiGGer and the optimised design produced by our new algorithm. Because DiGGer does not account for kinship information, its optimisation is not affected by this additional information. As a result, because the strong structure present in the initial design is not altered, the value of the optimisation function, accounting for kinship, is fairly poor in this case (1.16285241). On the other hand, we can see that the new optimization procedure incorporates the kinship information well by producing a check pattern where all three families are interwoven. This is confirmed by a well-improved optimization function (1.15030587) as shown by the algorithm convergence speed (convergence reached in $23.7$ seconds) in Fig.~\ref{fig:Phase II cv with kinship}. This highlights the importance, when using DiGGer, to randomize the families of genotypes within the field before running the optimization. Doing so on this example leads to a much improved value (1.15371275) of the optimisation function of the design produced by DiGGer compared to when the \textit{experimental} genotypes are not randomized, although our algorithm still betters it (Fig.~\ref{fig:Phase II cv with kinship}).

\section{Conclusion}\label{sec:conclusion}

Previous strategies of allocating genotypes between and within location in experimental designs such as the sparse and repchecks designs both ignore kinship information. When allocating the genotypes between locations, this was done at random (i.e. without necessary ensuring a good spread of families across locations) and, within location, this was done by only ensuring a good spread of the repeated \textit{check} genotypes (without accounting for the relatedness of the \textit{experimental} genotypes).

Through the case studies presented in this paper, we have shown that our new algorithm ensures that the kinship is well accounted for, therefore limiting the risks of confounding between locations and genotypes families. Within location, the gain from the current strategy is two-fold. First, we have shown that the convergence time from the existing algorithm is reduced (approx. 50\%) when kinship is not accounted for and that existing patterns of close \textit{check} genotypes are no more present. Second, accounting for kinship ensures that no more confounding is present within a location, even if a strong pattern of families is present in the starting design matrix.

This paper has concentrated on the repchecks design; however, it could easily be generalized to other designs by defining new design matrices and including additional, potentially complex, constraints. Such examples include uniqueness of some genotypes within location or, in the case of p-reps designs \cite{Williams2011b}, fixed proportions of duplicated genotypes within location. The same approach could be used to allocate entries to testers for topcross production and testing in the context of hybrid crop breeding. Further work is needed to incorporate these generalizations.

Although the examples shown in this paper were restricted to cases where the families of genotypes were simple (independent families of full siblings), the results presented can be extended, without loss of generality, to much more complex kinship matrices, e.g. based on molecular data, to ensure a more accurate definition of the relationships between genotypes.

This paper has concentrated on comparing the new approach to one existing approach, DiGGer. Work is ongoing to further compare it to the other widespread approach, OD, for which the objective function used is closer to that used in our approach than that used in DiGGer. However, early comparisons show a very clear benefit, in computation time for using our approach compared to OD, whose convergence times often make it impractical to use in real-life cases, with hundreds of genotypes and plots over several locations.

In summary, new ideas of Differential Evolution adaptation to combinatorial permutation-based optimization are presented in this paper. This is the first time Differential Evolution is applied for generating efficient experimental designs. Although this problem is hard to solve, good convergence properties of the new algorithm allow to find promising sparse designs for real large-size problems. The software is extremely optimized and tuned for the last generations of Intel processors (multi-threaded, vectorized, …) making the computations faster than existing software. This opens horizons for larger and more efficient designs in the future.


\bibliographystyle{IEEEtran}
\bibliography{optiLocation}
%\bibliography{../../biblio/nicolas,optiLocation,../../biblio/library}

\end{document}